\documentclass[runningheads]{llncs}
\usepackage[T1]{fontenc}
\usepackage{subcaption}
\usepackage{mathtools}
\usepackage{xcolor}
\newcommand\independent{\protect\mathpalette{\protect\independenT}{\perp}}
\def\independenT#1#2{\mathrel{\rlap{$#1#2$}\mkern2mu{#1#2}}}

\usepackage{mwe}
\usepackage[utf8]{inputenc}
\usepackage{amssymb}
\usepackage{graphicx}
\usepackage{mwe}
\usepackage{yhmath}
\usepackage{amsmath}
\usepackage{authblk}
\newcommand{\blue}{\textcolor[HTML]{1f77b4}}
\newcommand{\orange}{\textcolor[HTML]{ff7f0e}}
\newcommand{\green}{\textcolor[HTML]{2ca02c}}
\newcommand{\red}{\textcolor[HTML]{d62728}}
\newcommand{\purple}{\textcolor[HTML]{9467bd}}
\newcommand{\brown}{\textcolor[HTML]{8c564b}}
\newcommand{\pink}{\textcolor[HTML]{e377c2}}
\newcommand{\gray}{\textcolor[HTML]{7f7f7f}}
\newcommand{\chartreuse}{\textcolor[HTML]{bcbd22}}

\usepackage[symbol]{footmisc}

\title{Probabilistic Temporal Prediction of Continuous Disease Trajectories and Treatment Effects Using Neural SDEs}
\author{Joshua Durso-Finley$^{*,1,4}$, Berardino Barile$^{*,1,4}$,Jean-Pierre Falet$^{3,4}$, Douglas L. Arnold$^{1,3}$, Nick Pawlowski$^2$, Tal Arbel$^{1,4}$}

\authorrunning{Durso-Finley, J. and Barile, B}
\institute{$^1$Center for Intelligent Machines, McGill University. $^2$Microsoft Research. $^3$Montreal Neurological Institute, McGill University. $^4$MILA (Quebec AI institute).\\
$^{*}$ These authors contributed equally}
\begin{document}
\maketitle
\begin{abstract}

Personalized medicine based on medical images, including predicting future individualized clinical disease progression and treatment response, would have an enormous impact on healthcare and drug development, particularly for diseases (e.g. multiple sclerosis (MS)) with long term, complex, heterogeneous evolutions and no cure.  In this work, we present the first stochastic causal temporal framework to model the continuous temporal evolution of disease progression via Neural Stochastic Differential Equations (NSDE). The proposed causal inference model takes as input the patient's high dimensional images (MRI) and tabular data, and predicts both factual and counterfactual progression trajectories on different treatments in latent space. The NSDE permits the estimation of high-confidence personalized trajectories and treatment effects. Extensive experiments were performed on a large, multi-centre, proprietary dataset of patient 3D MRI and clinical data acquired during several randomized clinical trials for MS treatments. Our results present the first successful uncertainty-based causal Deep Learning (DL) model to: (a) accurately predict future patient MS disability evolution (e.g. EDSS) and treatment effects leveraging baseline MRI, and (b) permit the discovery of subgroups of patients for which the model has high confidence in their response to treatment even in clinical trials which did not reach their clinical endpoints.
\end{abstract}

\section{Introduction}
\label{sec:intro}
Consider patients with long-term, incurable disease, characterised by complex and heterogeneous disease courses and treatment responsiveness.  In multiple sclerosis (MS), a common disabling neurological condition, no strongly predictive imaging marker exists for progressive disability evolution \cite{MSheterogeneity}, and it's unclear if available drugs could be effective at slowing disability progression in subsets of patients, as their efficacy at the population level has been limited. In these contexts, the impact of a deep learning (DL) clinical decision support tool that can predict the future continuous disease evolution for individual patients based on (non-invasive) medical scans, and standard clinical and demographic information, would be enormous. Even more impactful would be if it could predict the relative effect of available treatments compared to an untreated course for \textit{this patient}, and quantify the degree of confidence in its predictions for safe deployment in high-risk settings. Such is the promise of DL for image-based precision medicine. Unfortunately, this promise has not yet been met \cite{AIMSsurvey}.

Predictions of clinical disease outcomes have, for the most part, focused on outcomes at fixed future timepoints\cite{tousignant2019,FANGCovid,litjens2017survey}. In MS, some have proposed to model particular radiomic feature evolutions and link them to clinical outcomes (\cite{pontillo2022,eshaghi2021}). Learning the dynamics of long-term disease evolution for \textit{clinical} endpoints based on medical images would be hugely beneficial to patient care in general. This would lead to a better understanding of disease evolution across the population, including the discovery of new personalized predictive image markers, thereby providing the basis for more equitable patient care. Furthermore, forecasting continuous future clinical outcome trajectories based on early observations, especially with irregularly spaced, missing and noisy observations, would be enormously impactful for improved patient care and clinical trial analysis.
 
Recently, Neural Ordinary Differential Equations (NODEs) \cite{Chen2018} were proposed to improve upon discrete-time models based on RNNs \cite{SerstinskyRNN}. NODEs were applied to forecasting in clinical settings but only with tabular or synthetic input data \cite{NeuralODEpharmakinetics,Tang2022,DeBrouwer2019,Norcliffe2023}. Given that measured clinical outcomes can be noisy (e.g. Expanded Disability Status Scale \cite{UncertaintyEDSS} in MS), and that DL models are not always accurate, probabilistic temporal models that represent a distribution over possible trajectories would be desirable and help quantify the uncertainties in the model's predictions. Although many approaches for estimating uncertainty have been proposed for DL predictions in medical imaging~\cite{UncertaintySurvey,BayesianMedicineSurvey}, applications in personalized medicine have been restricted to single time-point predictions \cite{dursofinley2023improving}. Neural stochastic differential equations (NSDEs)\cite{Xuechen2020} have recently been proposed as an extension of NODE for probabilistic modeling of trajectories in latent space for tabular data \cite{VDSintegratingNODE}. However, to our knowledge, NSDE-based methods have not been applied to medical image analysis. Finally, in chronic conditions where many treatments are available, predicting a future individual treatment effect (ITE) requires estimating the relative effect of one treatment against a comparator. Causal inference for prediction of future ITE in medical image analysis remains in its infancy \cite{SotosCMLhealthcare}, with some recent work focused on future fixed-timepoint outcomes\cite{dursofinley2022personalized,ma2023treatmentHemmorage}. The importance of causality for image-based personalised medicine warrants extending prior work on causal modeling for ITE estimation to NSDEs, which have not been studied in this context.

In this paper, we present the first probabilistic temporal causal model for image-based personalized medicine. The encoder combines the pre-treatment (baseline) high-dimensional images (i.e. MRI) with tabular clinical data, and projects its latent representation forward in time to model the continuous evolution of clinical trajectories via an NSDE. The model infers ITE by comparing the predicted factual and counterfactual future clinical disease trajectories for patients on all available treatments in latent space and additionally uses the stochastic nature of the NSDE for sampling-based variance estimation to quantify uncertainty in the predicted trajectory (and the ITE on the trajectory). Extensive experiments are performed on a large (over 3600 patients), multi-centre, proprietary dataset of 3D MRI acquired from patients with MS during six randomized clinical trials (RCTs), testing seven active treatments, and two placebo groups. We illustrate results on two different subtypes of MS: relapsing remitting MS (RRMS) and secondary progressive (SPMS). Compared to a discrete temporal model (LSTM) and a fixed time-point Encoder-Decoder regressor (see Supplemental Materials, Figure 2 for details), our model more accurately predicts future EDSS. Moreover, we find that sampling-based uncertainty estimates allow for isolating subsets of patients for which the model is more confident, improving the predictive accuracy of the trajectory and ITE.

\section{Method}
\label{sec:method}
\begin{figure}[h]
\label{fig:framework}
\centering
\includegraphics[width=0.8\textwidth]{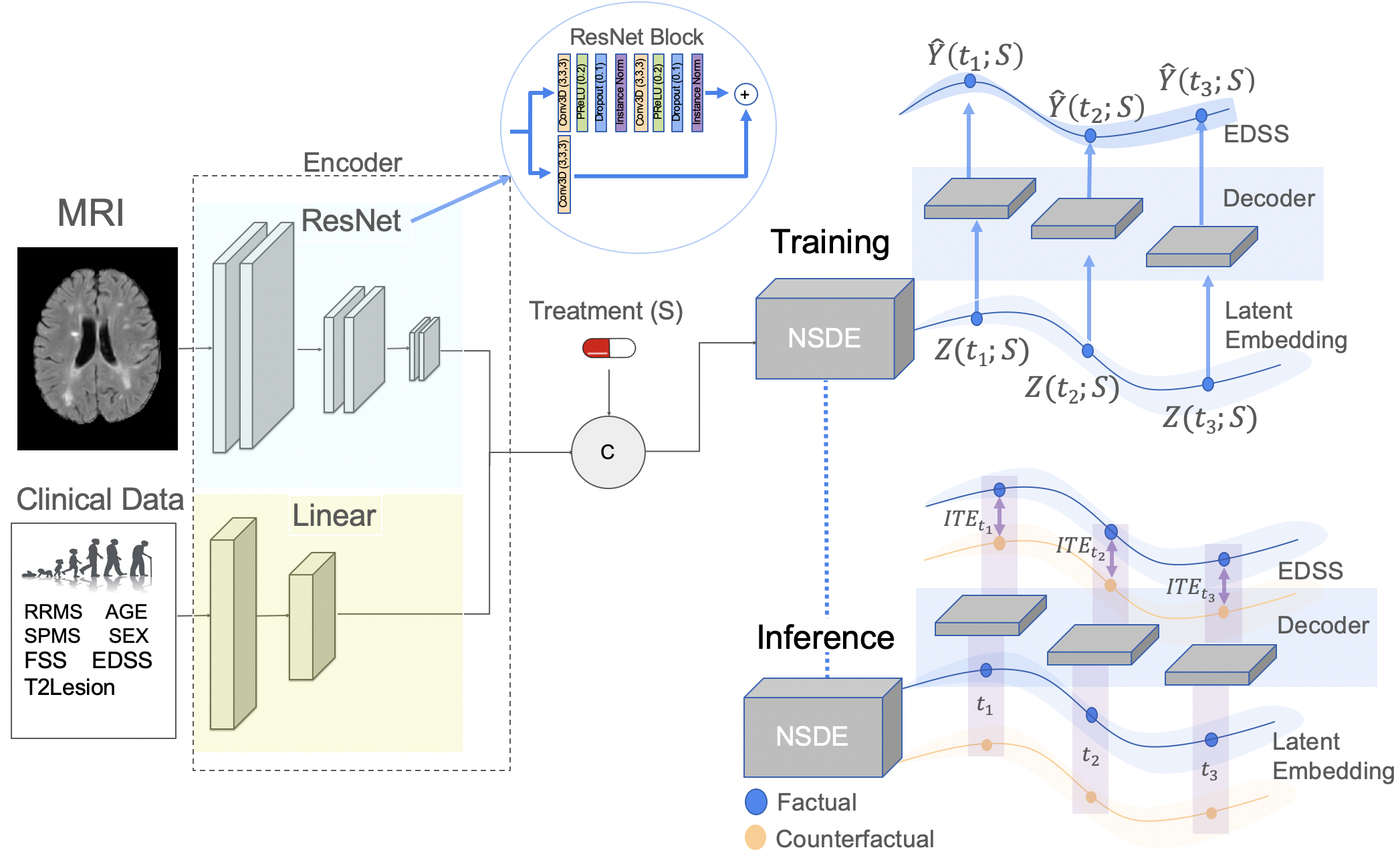}
\caption{Overall framework: 3D MRI and tabular data (clinical, demographic, and subtypes) are encoded, concatenated and passed to the NSDE solver. During supervised \textbf{Training} (end-to-end), the model learns to represent the temporal trajectory in latent space conditioned on the treatment the patient received. The decoder maps the predicted longitudinal latent representations into outcomes (EDSS scores). During \textbf{Inference}, the model projects the embeddings onto the latent space forwarded in time and conditioned on the treatment (or control), permitting factual and counterfactual latent trajectories and uncertainties to be estimated. The decoder predicts future outcomes, and uncertainties, at any point along the continuous latent trajectories. ITE, and associated uncertainties, can be estimated by comparing probabilistic trajectories on and off treatments.}
\label{fig:diagram}
\end{figure}

A causal stochastic temporal model is proposed: At baseline, multi-dimensional MRI, and the associated tabular data (clinical, demographic, and subtype), are encoded in two distinct lower-dimensional latent spaces. The treatment label, one-hot-encoded, is concatenated together with the tabular and image embeddings and fed to the NSDE which projects the combined latent representation (from baseline) forward in time. Finally, a decoder is trained to map the predicted latent encoding to the outcome of interest (e.g. future EDSS) (see Figure \ref{fig:diagram}). The general mathematical framework is now described.

\subsection{Learning Probabilistic Trajectories via Stochastic Neural ODE}
Let $t \in \mathbb{R}$ be time where $t_0$ represents the time of the initial visit. Let $X(t_0) = \{ x_i(t_0) \in \mathbb{R}^{(c_1,c_2,c_3)} \}_{i=1}^n$ be the 3D MRI at time $t=0$, $i$ the patient index, $(c_1,c_2,c_3)$ the image size and $W(t_0) = \{ w_i(t_0) \in \mathbb{R}^{d} \}_{i=1}^n$ be the associated clinical data. $X(t_0)$ and $W(t_0)$ are mapped to a lower dimensional representation $Z_x(t_0) = \phi(X(t_0);\theta_\phi) \in \mathbb{R}^{h_1}$ and $Z_w(t_0) = \psi(W(t_0);\theta_\psi) \in \mathbb{R}^{h_2}$, where $\phi$ and $\psi$ are trainable functions parametrized by $\theta$. Let $Z(t_0)$ be the combined latent representation obtained by concatenating the two latent vectors such that $Z(t_0) \in \mathbb{R}^{h_1+h_2}$. For each time series, we want to predict the clinical target $Y_i(t \mid S=s)$ at some timepoint $t > t_0$ in the future condition to treatment $S=s$. We use the NSDE to model a stochastic process for Z whose dynamics are derived from a deterministic (drift) function $f_\theta$ and a stochastic (diffusion) component $g_\theta$:
 \begin{equation}
    \frac{dZ(t)}{dt} = f_\theta\left(Z(t) \mid S=s\right) + g_\theta \left(Z(t) \mid S=s\right) \circ d\Omega_t
\end{equation}
\noindent where $d\Omega_t \sim \mathcal{N}(0,\Delta t I_\omega)$ represents the $h_1+h_2$-dimensional Brownian motion. In our model, $f_\theta$ and $g_\theta$ are parameterized by neural networks. The latent encoding for a future latent state given treatment $s$ and initial state $Z(t_0)$ can be obtained through Stratonovich integration \cite{Kidger2021} as follows:
\begin{equation}
    Z_{i,j}(t) = \text{ODESolve}(Z_{i}(t_{0}),f_\theta(\cdot) + g_\theta(\cdot) \circ \frac{d\Omega_{t,j}}{dt} \mid S=s, [t_0,t]).
\end{equation}
\noindent Thus, $Z_{i,j}(t)$ represents the forecasted latent space for patient “$i$”, with Brownian motion sample “$j$”, at future time “$t$”, and conditioned on the treatment assignment $S=s$. Lastly, the latent state is decoded through a deterministic function parameterized by a neural network $\Phi(Z_{ij}(t) \mid S=s ; \theta_{\Phi}) \in \mathbb{R}$ to obtain predicted outcome $\hat{Y}_{ij}(t \mid S=s)$.

\subsection{Temporal Counterfactual Trajectories and ITE}
\label{sec:temp_potential}
In our proposed framework, we extend individual treatment effect (ITE) \cite{shalit2017TARNET} estimation to the longitudinal setting, by comparing latent \textit{trajectories} of patients on and off treatments from baseline inputs. Following the Neyman/Rubin Potential Outcome Framework \cite{Imbens2015}, we consider two potential outcomes $y_0$ (control/placebo) and $y_s$ (treatment $s$). For patient $i$ with triples $\mathcal{D} = \{(z_{i,t}, y_{i,t}, s_{i,t})\}_{i=1}^n$ where $t>t_0$, the ITE can be defined as a function of time:  $\text{ITE}_s(t)=y_s(t) - y_0(t)$, where $y_s(t)$ and $y_0(t)$ represent \textit{potential} outcomes for treatment and control patients, respectively.
The $\text{ITE}_s(t)$ is an unobservable causal estimand, as only one of the two potential outcomes can be observed. For each patient $i$ and Brownian sample $j$, the following causal estimand for ITE, conditioned on the latent encoding $Z$ at timepoint $t$, is used: $\tau_{s}^{(ij)}(z;t) = \mathbb{E}[y_s \mid z;t]_{ij} - \mathbb{E}[y_0\mid z;t]_{ij}$.
In the case of RCTs (our case), $(Y_0, Y_s) \independent S \mid Z(t_0)$, and therefore we can learn an unbiased estimator for  $\tau_{s}$ \cite{PearlDoCalculus}. A trajectory ITE can thus be estimated as follows: 
\begin{equation}
   \bar{\tau}_{s}^{(ij)} = \frac{1}{t_k-t_0} \sum_{\forall t \geq t_0} \tau_{s}^{(ij)}(z;t) 
   \label{eq:averageITE}
\end{equation}
where $k$ is the number of longitudinal timepoints. As the NSDE permits stochastic sampling of predicted factual and counterfactual trajectories, we can integrate over $j$ and obtain the mean, $\vec{\mu}^{(i)}_s(z)$, and variance, ${\vec{\sigma}}^{(i)}_s(z)$, of the longitudinal ITE for patient $i$ (Eq. (\ref{eq:averageITE})).
The variance can thus be used as an uncertainty estimand for the longitudinal ITE.

\section{Experiments and Results}
\label{sec:expts}

\subsection{Dataset}
The data is pooled from six randomized clinical trials (RCTs)\footnote{Specific references to papers describing the details about the curated dataset cannot be provided at submission without breaking anonymization.}. Five of them,  BRAVO \cite{BRAVO}, OPERA 1 \cite{OPERA}, OPERA 2 \cite{OPERA}, DEFINE \cite{DEFINE}, and ADVANCE \cite{ADVANCE_ATTAIN} enrolled patients with relapsing-remitting MS (RRMS) and randomized them to one of the following treatments: \blue{Placebo} ($n=455$), \orange{Laquinimod} ($n=295$), \green{INterFeron Beta-1a IntraMuscular (INFB-IM)}  ($n=326 $), \purple{INterFeron Beta-1a SubCutaneous (INFB-SC)} ($n=572$), \brown{DiMethyl Fumarate (DMF)}  ($n=225$),  \red{Ocrelizumab} ($n=554$), \pink{Pegylagted-Interferon (PINFB)} (n=840). ASCEND \cite{ASCEND} randomized secondary progressive MS (SPMS) patients to \chartreuse{Placebo} ($n=302$) and \gray{Natalizumab} ($n=293$) (See Supplemental Material, Table 1 for more dataset statics.)  The primary clinical target of interest is the EDSS score, which measures rate of disability. It is an ordinal rating scale ranging from 0 (no disability) to 10 (death) in steps of 0.5. Inputs to the model consisted of: (1) clinical and demographic features available at baseline and pre-trial screening visit: age, sex, functional scores (FSS), T2 lesion volume and EDSS. (2) Fluid Attenuated Inverse Recovery (FLAIR) MRI acquired at baseline at $1 \times 1 \times 3$ mm resolution. Follow-up EDSS scores, measured inconsistently over time at intervals of approximately 12 weeks, up until 96 weeks, were used as regression targets.

\subsection{Experimental and Implementation Details}
A ResNet encoder~\cite{He2016} with 3 residual blocks, and a feed-forward network with 2 linear layers and ELU activations, were used to encode the 3D MRI and the associated clinical data, respectively. Balanced MSE \cite{BalancedMSE} was used as the regression loss. The entire model was trained end-to-end and evaluated as part of a nested $4 \times 4$-fold Cross-Validation (CV) strategy. Specifically, the dataset was initially split into 4 folds. At each iteration, each fold was used as a held-out test set, while the remaining 3 folds were used for training, validation and hyperparameter tuning. The process was repeated until each of the 4-folds was used as hold-out test set. Results are presented based on CV aggregation or crogging \cite{Crogging}, to improve the generalization error estimate. During training the model learns from observed factual outcomes. During inference, estimates for the counterfactual trajectories and ITE are obtained by sampling trajectories conditioned on specific treatments.
\subsection{Evaluating Predicted EDSS Trajectories and Uncertainties}
\label{sec:Factual}
\begin{figure}
\centering
\begin{subfigure}{.5\textwidth}
  \centering
    \includegraphics[width=\linewidth]{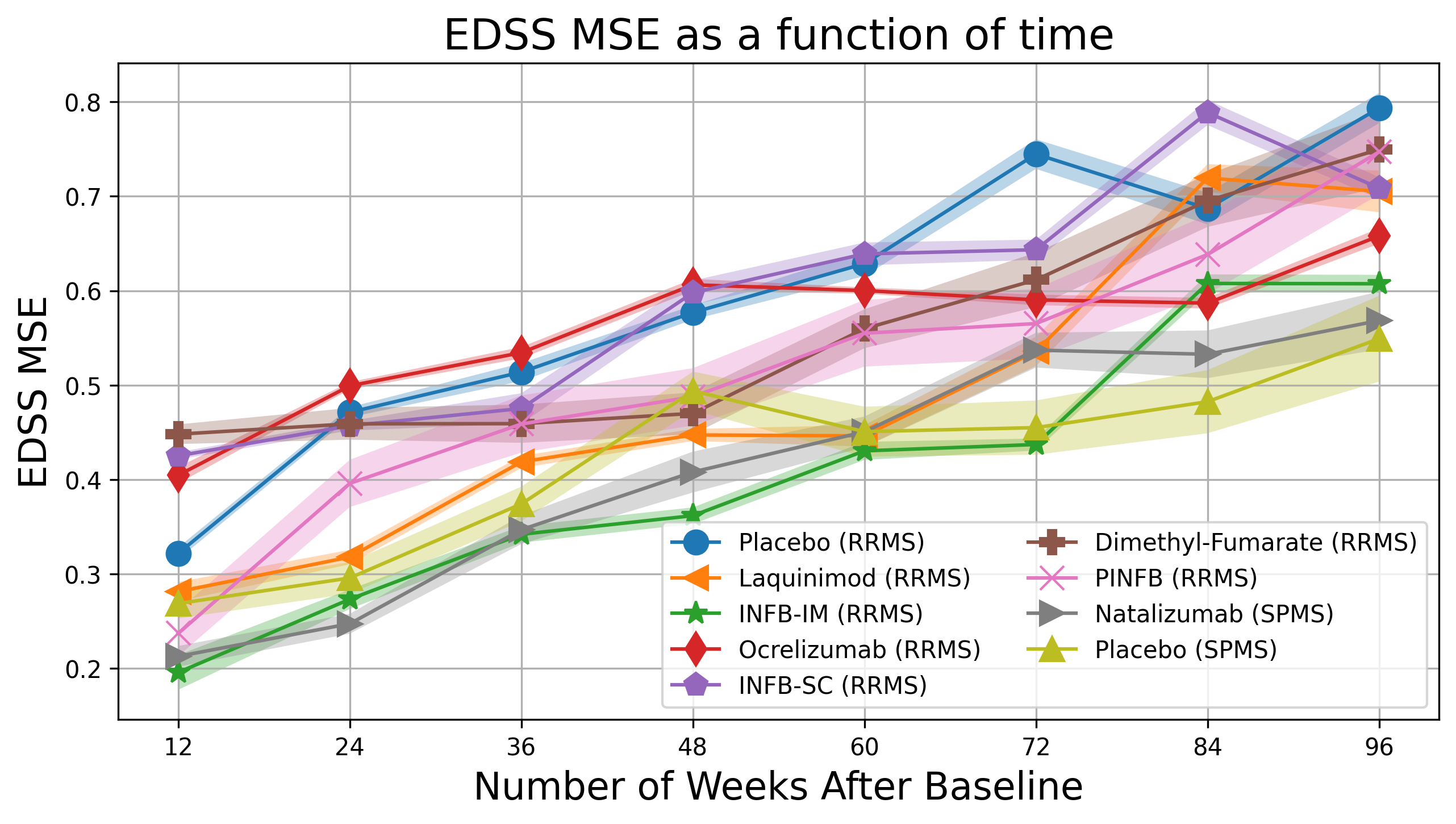}
    \caption{}
  \label{fig:msea}
\end{subfigure}%
\begin{subfigure}{.5\textwidth}
  \centering
     \includegraphics[width=\linewidth]{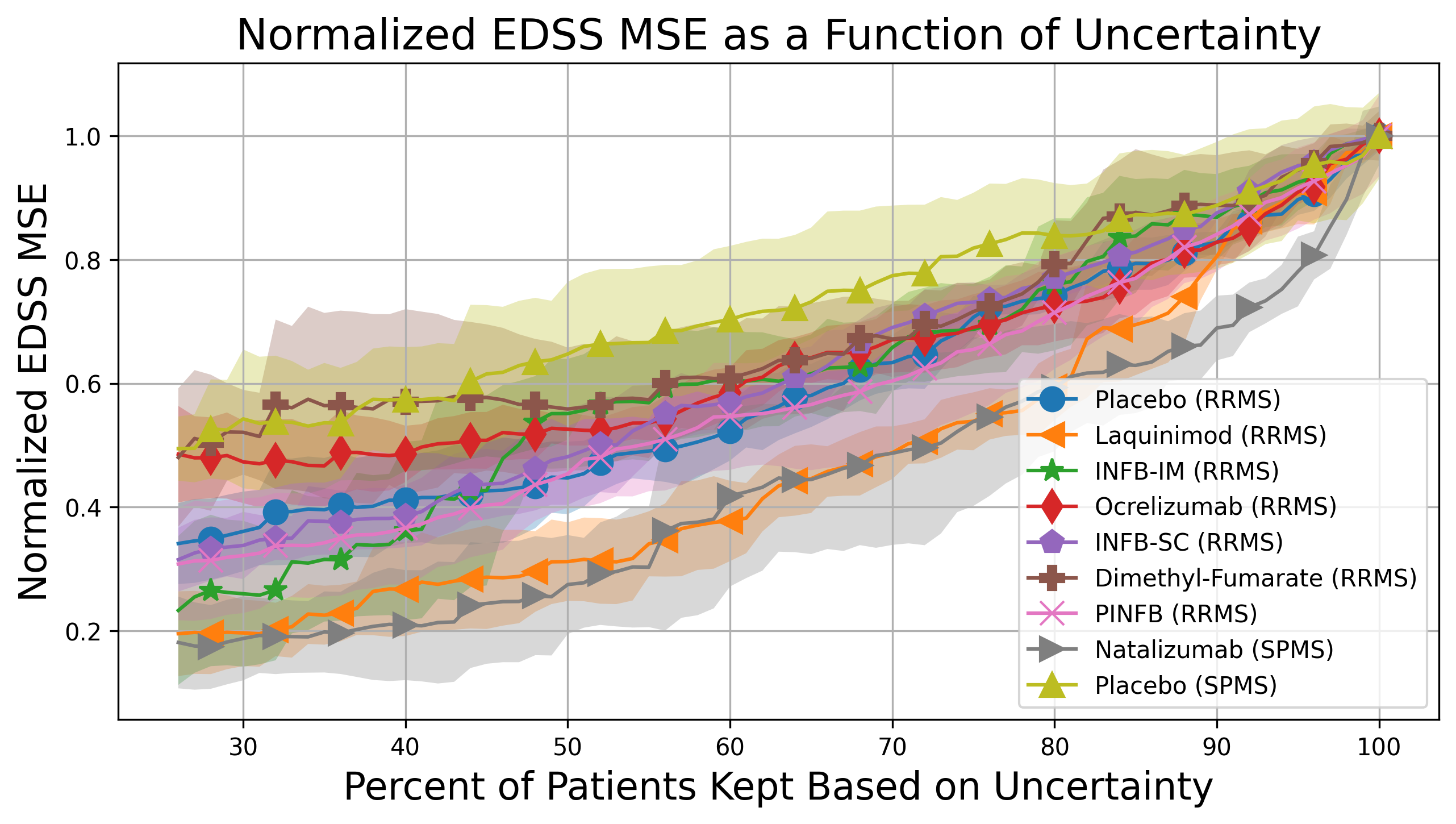} 
  \caption{}
  \label{fig:mseb}
\end{subfigure}
    \caption{(a) MSE of factual EDSS predictions as a function of time. Evaluations are shown at  12-week intervals up to week 96. Results are shown for 7 active treatments and 2 placebo control groups (RRMS and SPMS). (b) Normalized Mean Squared Error (MSE) of factual EDSS predictions for patients kept based on model uncertainties. One can see a clear decrease in error when moving from right (100\% of the patients) to left (30\% of the patients with the most confident predictions). Shaded areas show the variance from different sets of outer fold aggregation.}
\label{fig:mse_results}
\end{figure}

Evaluations on the factual predictions for future EDSS scores are performed. Fig.\ref{fig:mse_results}(a) reports the MSE between available and predicted EDSS scores. These are provided at (approximately) 12-week intervals up until week 96. To estimate uncertainty, the NSDE was run 30 times with different Brownian noise samples for each prediction. For all patients, the MSE was below 1 at all timepoints, demonstrating that the model accurately predicts disease evolution up to 2 years in the future (e.g. ~\cite{Norcliffe2023} for comparison). As expected, more accurate results were obtained earlier in the trajectory, with reduced performance farther into the future. 

\noindent The proposed approach was compared with different baselines: (1) A Long-Short-Term-Memory (LSTM) model was trained, where irregular visits were grouped together to the closest 12-week interval and (2) A fixed time-point Encoder-Decoder regressor. (The architectural details of the models are found in the Supplemental Materials.) Results are found in Table \ref{tab:factualMSEtableTransposed}. The proposed NSDE framework had an overall MSE of 0.575 (0.012), as compared to 0.674 (0.014) (non-temporal) and 0.662 (0.014) (LSTM) averaged over all treatments. 

\begin{table}
    \centering
    \begin{tabular}[width=\linewidth]{|c|c|c|c|}
             \hline
          Treatment  &  Encoder-Decoder & LSTM & NSDE \\ 
          \hline
         Laquinimod & 0.769 (0.022) & 0.691 (0.006) & \textbf{0.672} (0.008) \\
         INFB-SC     & 0.624 (0.004) & 0.614 (0.010) & \textbf{0.590} (0.006)  \\
         Ocrelizumab & 0.603 (0.008) & 0.616 (0.012) & \textbf{0.555} (0.003) \\
         INFB-IM     & 0.737 (0.009) & 0.712 (0.010)  & \textbf{0.699} (0.010) \\
         Dimethyl Fumarate & 0.708 (0.015) & 0.690 (0.005) & \textbf{0.679} (0.011)  \\
         PINFB           & 0.572 (0.005) & 0.584 (0.004) & \textbf{0.522} (0.008) \\
         Natalizumab     & 0.882 (0.024) & 0.857 (0.015) & \textbf{0.533} (0.013) \\
         Placebo (SPMS)  & 0.752 (0.013) & 0.675 (0.021) & \textbf{0.358} (0.013)  \\
         Placebo (RRMS)  & 0.672 (0.006) & 0.696 (0.008) & \textbf{0.655} (0.006)  \\
          \hline
     \end{tabular}
     \caption{Comparison of Factual Results across Baselines: Overall Mean and standard error, computed as the  average MSE for the EDSS outcomes across timepoints, and across patients, for three models: (1) A fixed time-point Encoder-Decoder regressor, (2) LSTM, and (3) proposed NSDE. 
     The proposed NSDE model generally outperforms the other models in all drugs and placebo groups. See Supplemental Materials for model details.}
     \label{tab:factualMSEtableTransposed}
 \end{table}

\noindent Evaluation of the uncertainty estimates is performed in order to confirm that higher confidence estimates correlate with lower errors. To this end, patients were ordered based on the model confidence (see Section~\ref{sec:temp_potential}). Fig.\ref{fig:mse_results} reports the MSE results averaged along the entire trajectories for patients filtered based on their uncertainty estimates. To better analyze the impact of uncertainty on error, the MSE was normalized to the baseline case where no uncertainty filtering was performed (100\% of the data). Results indicate that trajectory predictions for all treatments decrease in error when more confident predictions are considered, confirming the desired performance of the predicted uncertainties.  For example. when considering the 30\% most confident predictions (x-axis), the normalized MSE score was at 0.18 for Natalizumab, which was 83\% lower than the reference case (no-uncertainty).

\subsection{Counterfactual Predictions and Treatment Response}
\label{sec:CDP}
\begin{figure}

    \begin{subfigure}{0.33\textwidth}
            \includegraphics[width=\textwidth]{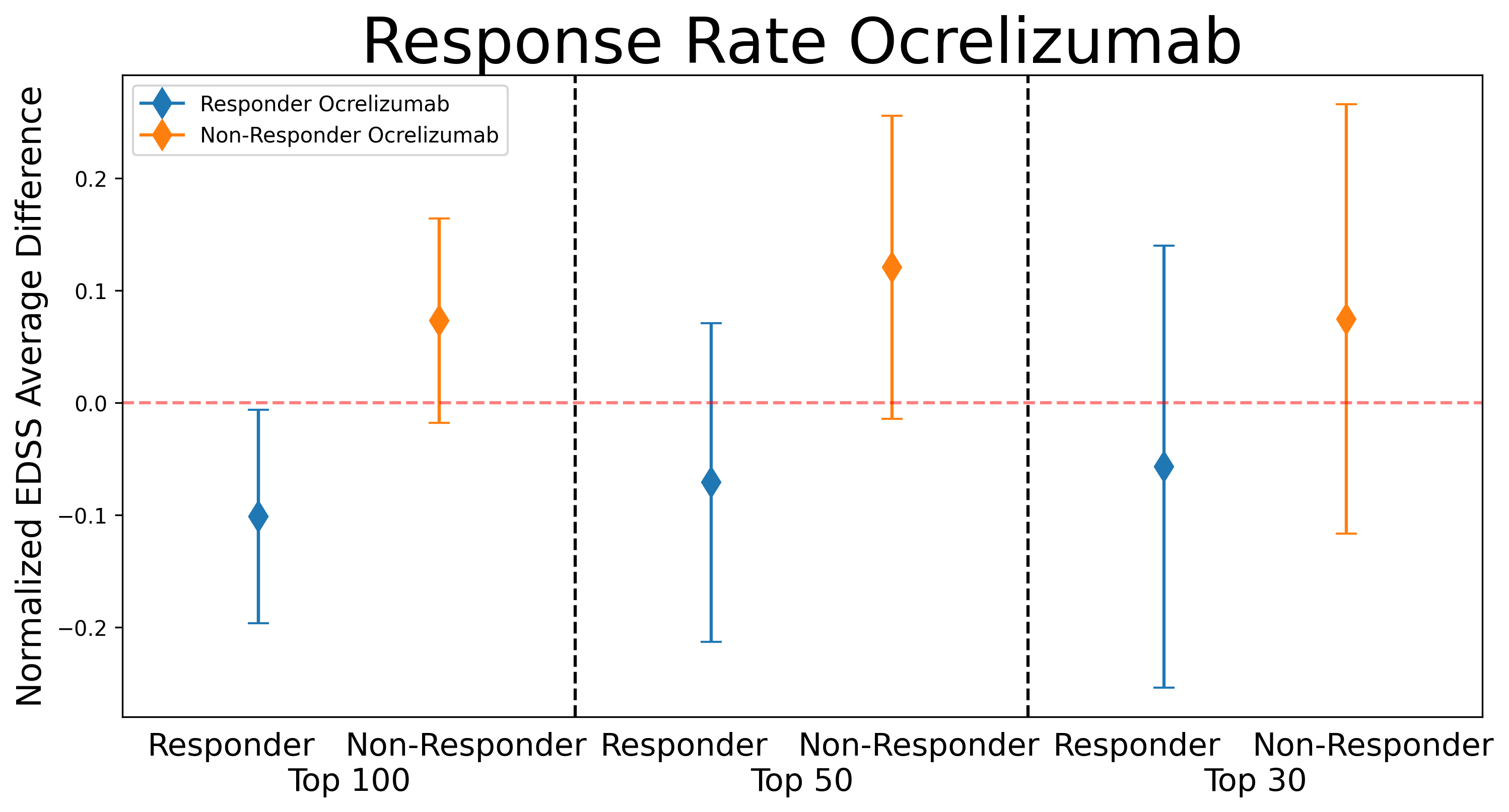}   
            \caption{} 
            \label{fig:ResponseGroupsA}
  \end{subfigure}\hfil
    \begin{subfigure}{0.33\textwidth}
            \includegraphics[width=\textwidth]{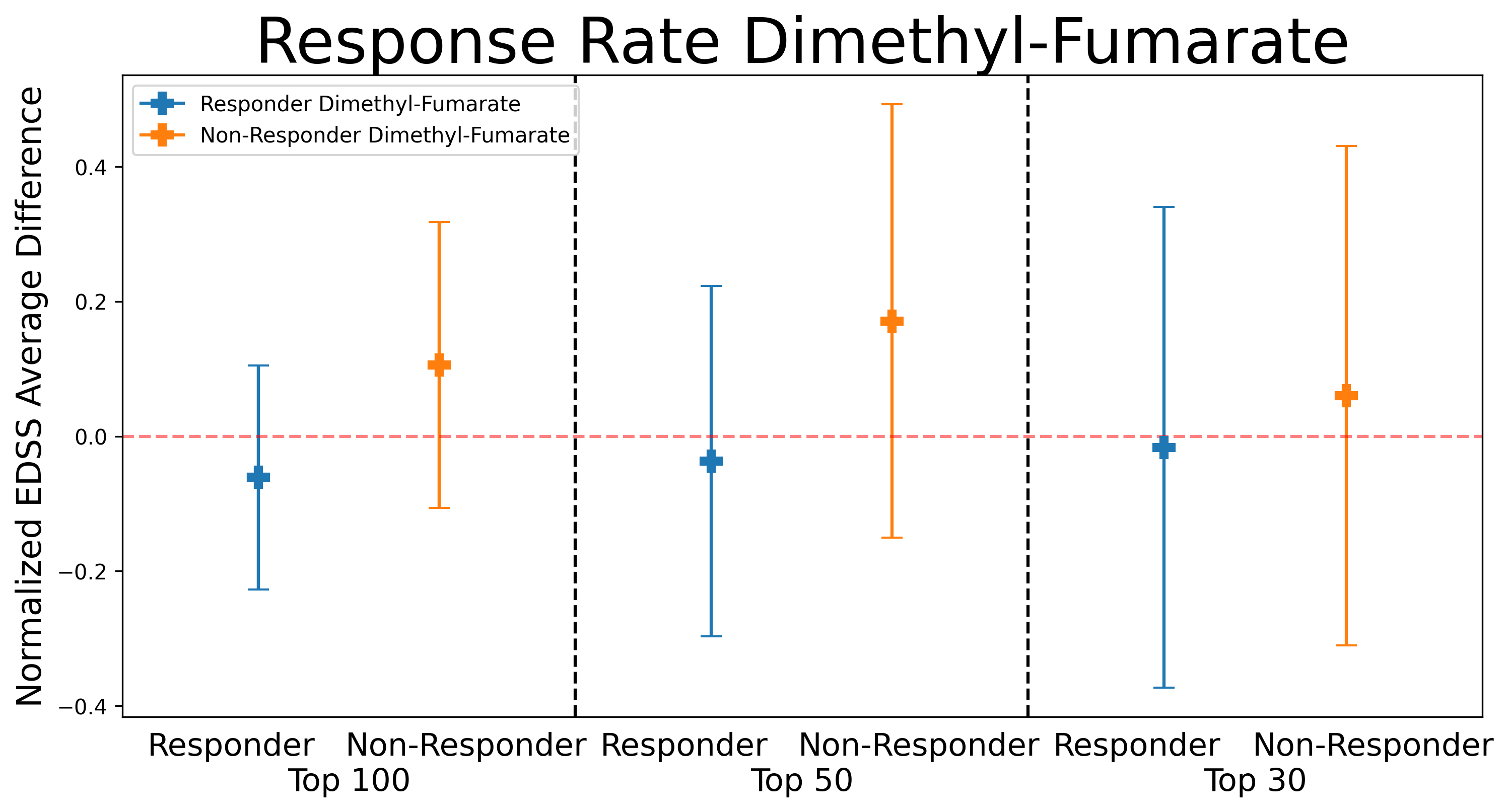}   
            \caption{} 
            \label{fig:ResponseGroupsB}
  \end{subfigure}\hfil
      \begin{subfigure}{0.33\textwidth}
            \includegraphics[width=\textwidth]{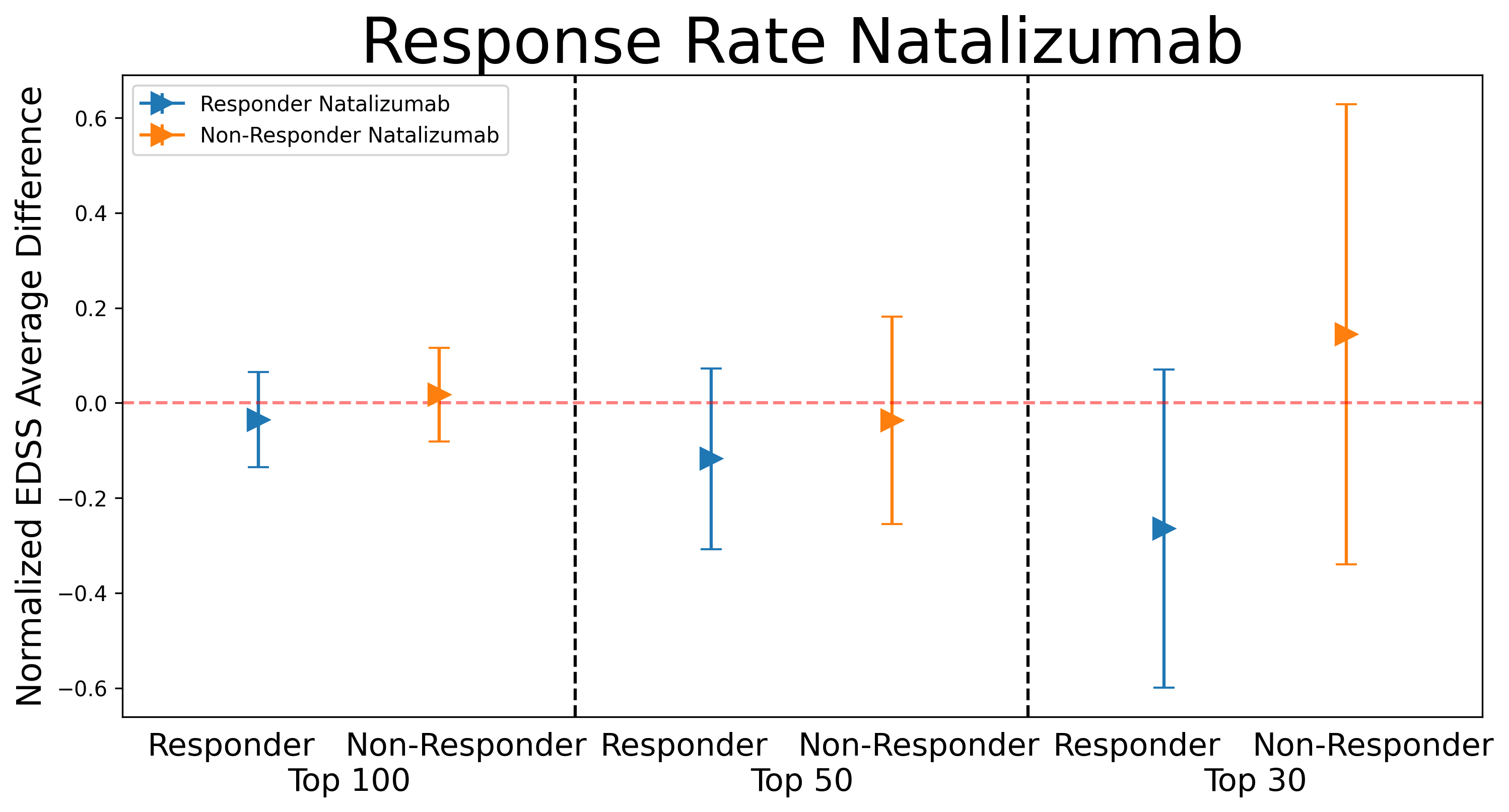}   
            \caption{} 
            \label{fig:ResponseGroupsC}
  \end{subfigure}\hfil
    \caption{Treatment response estimates (uplift) for three drugs (Ocrelizumab, DMF, Natalizumab). Comparison of predicted \blue{responders} and \orange{non-responders} at three different levels of uncertainties (30\%, 50\% and 100\%-no uncertainty). Negative values indicate response through a reduction in average EDSS change on the drug as compared to placebo.  Our model finds subgroups of responders for these drugs, known to be of moderate and high efficacy. The others drugs are considered to be low efficacy (Results in Supplemental Materials).}
    \label{fig:ResponseGroups}
\end{figure}

The proposed model predicts future individual counterfactual (CF) trajectories of patient disability scores (EDSS), resulting from a causal intervention on the treatment variable. Given that validating CFs is challenging in the absence of ground truth, evaluations are based on \textit{uplift} scores~\cite{Verhelst2023}, defined as the average difference in predicted EDSS change between treated and counterfactual (non-treated) patients. When sorting patients by their uplift score, predicted \blue{responders}, and \orange{non-responders} can be identified. Responders are expected to have negative uplift values, indicating that the change in EDSS is higher on placebo than it is on treatment. Figure~\ref{fig:ResponseGroups} shows 3 examples of uplift plots for SPMS patients treated with Natalizumab, and RRMS patients treated with either DMF or Ocrelizumab. These were chosen because they are known to be moderate and high efficacy drugs\cite{Montalban2017,DEFINEextendedstudy}. Results are depicted for three levels of uncertainties in order to explore whether the identification of responders to treatment can be further improved by considering the subset of patients with high predicted confidence. The results indicate that uplift scores are generally negative (lower) for the responder group (33\% of patients with highest uplift) compared to the non-responder (33\% of patients with lowest uplift), suggesting that the model is able to identify responders with high degree of accuracy. When comparing the 30\% of patients with the highest confidence predictions, the gap between responders and non-responder widens, suggesting an even greater treatment effect for these patients.

\section{Conclusions}
For chronic neurological diseases such as MS, the ability to accurately predict continuous individual trajectories of disease evolution on different treatments from baseline MRI would be of substantial importance to improving patient care and drug development. In this work, a probabilistic temporal model is proposed for (1) predicting continuous disability trajectories in latent space based on NSDEs, and (2)  future individual treatment responses, and validated on a large (>3600 patients) unified RCT dataset for a variety of treatments. Causal inference, combined with uncertainty estimates, provides additional trustworthiness in predicted counterfactual trajectories, and permit the identification of high confidence subgroups of highly responsive patients. Future work will explore other strategies to measure treatment response and uncertainty estimation over trajectories. Finally, clustering of subgroups of responders will permit the discovery of common predictive images markers across the population.

\newpage
\bibliographystyle{splncs04}
\bibliography{bibliography}

\begin{thebibliography}{10}
\providecommand{\url}[1]{\texttt{#1}}
\providecommand{\urlprefix}{URL }
\providecommand{\doi}[1]{https://doi.org/#1}

\bibitem{UncertaintySurvey}
Abdar, M., et~al.: A review of uncertainty quantification in deep learning: Techniques, applications and challenges. Information Fusion  (2021)

\bibitem{BayesianMedicineSurvey}
Ashby, D.: Bayesian statistics in medicine: A 25 year review. Statistics in medicine  (2006)

\bibitem{Crogging}
Barrow, D.K., Crone, S.F.: Crogging (cross-validation aggregation) for forecasting — a novel algorithm of neural network ensembles on time series subsamples. In: The 2013 International Joint Conference on Neural Networks (IJCNN) (2013)

\bibitem{DEFINE}
Bomprezzi, R.: Dimethyl fumarate in the treatment of relapsing–remitting multiple sclerosis: an overview. Therapeutic Advances in Neurological Disorders  (2015)

\bibitem{ADVANCE_ATTAIN}
Calabresi, P.A., et~al.: Pegylated interferon beta-1a for relapsing-remitting multiple sclerosis ({ADVANCE}): a randomised, phase 3, double-blind study. The Lancet Neurology  (2014)

\bibitem{Chen2018}
Chen, T., et~al.: Neural ordinary differential equations. CoRR  (2018)

\bibitem{DeBrouwer2019}
De~Brouwer, E., et~al.: Gru-ode-bayes: Continuous modeling of sporadically-observed time series. Advances in Neural Information Processing Systems  (2019)

\bibitem{dursofinley2023improving}
Durso-Finley, J., Falet, J.P., Mehta, R., Arnold, D.L., Pawlowski, N., Arbel, T.: Improving image-based precision medicine with uncertainty-aware causal models (2023)

\bibitem{dursofinley2022personalized}
Durso-Finley, J., Falet, J.P.R., Nichyporuk, B., Arnold, D.L., Arbel, T.: {Personalized Prediction of Future Lesion Activity and Treatment Effect in Multiple Sclerosis from Baseline MRI} (2022)

\bibitem{eshaghi2021}
{Eshaghi}, A., et~al.: {Identifying multiple sclerosis subtypes using unsupervised machine learning and MRI data}. Nature Communications  (2021)

\bibitem{FANGCovid}
Fang, C., et~al.: Deep learning for predicting covid-19 malignant progression. Medical Image Analysis  (2021)

\bibitem{DEFINEextendedstudy}
Gold, R., et~al.: Safety and efficacy of delayed-release dimethyl fumarate in patients with relapsing-remitting multiple sclerosis: 9 years' follow-up of define, confirm, and endorse. her Adv Neurol Disord  (2020)

\bibitem{OPERA}
Hauser, S.L., et~al.: Ocrelizumab versus interferon beta-1a in relapsing multiple sclerosis. New England Journal of Medicine  (2017)

\bibitem{He2016}
He, K., et~al.: Deep residual learning for image recognition. IEEE Conference on Computer Vision and Pattern Recognition (CVPR)  (2016)

\bibitem{Imbens2015}
Imbens, G., Rubin, D.: Causal inference for statistics, social, and biomedical sciences. Cambridge University Press  (2015)

\bibitem{ASCEND}
Kapoor, R., et~al.: Effect of natalizumab on disease progression in secondary progressive multiple sclerosis ({ASCEND}): A phase 3, randomised, double-blind, placebo-controlled trial with an open-label extension. The Lancet Neurology  (2018)

\bibitem{Kidger2021}
Kidger, P., et~al.: {Neural SDEs as Infinite-Dimensional GANs}. International Conference on Machine Learning  (2021)

\bibitem{litjens2017survey}
Litjens, G., et~al.: A survey on deep learning in medical image analysis. Medical image analysis  (2017)

\bibitem{NeuralODEpharmakinetics}
Lu, J., et~al.: {Neural-ODE for pharmacokinetics modeling and its advantage to alternative machine learning models in predicting new dosing regimens}. iScience  (2021)

\bibitem{MSheterogeneity}
Lucchinetti, C., et~al.: Heterogeneity of multiple sclerosis lesions: Implications for the pathogenesis of demyelination. Annals of neurology  (2000)

\bibitem{Montalban2017}
Montalban, X., et~al.: Ocrelizumab versus placebo in primary progressive multiple sclerosis. N Engl J Med.  (2017)

\bibitem{AIMSsurvey}
Naji, Y., et~al.: Artificial intelligence and multiple sclerosis: Up-to-date review. Cureus  (2023)

\bibitem{Norcliffe2023}
Norcliffe, A., et~al.: Benchmarking continuous time models for predicting multiple sclerosis progression. Transactions on Machine Learning Research  (2023)

\bibitem{PearlDoCalculus}
Pearl, J.: The foundations of causal inference. Sociological Methodology  (2010)

\bibitem{pontillo2022}
Pontillo, G., et~al.: Stratification of multiple sclerosis patients using unsupervised machine learning: a single-visit mri-driven approach. European Radiology  (2022)

\bibitem{VDSintegratingNODE}
Qian, Z., et~al.: Integrating expert {ODE}s into neural {ODE}s: Pharmacology and disease progression. In: Advances in Neural Information Processing Systems (2021)

\bibitem{BalancedMSE}
Ren, J., et~al.: Balanced {MSE} for imbalanced visual regression. In: IEEE Conference on Computer Vision and Pattern Recognition (CVPR) (2022)

\bibitem{UncertaintyEDSS}
Rudick, R.A., et~al.: {Disability Progression in a Clinical Trial of Relapsing-Remitting Multiple Sclerosis: Eight-Year Follow-up}. Archives of Neurology  (11 2010)

\bibitem{SotosCMLhealthcare}
Sanchez, P., et~al.: Causal machine learning for healthcare and precision medicine. Royal Society Open Science  (2022)

\bibitem{shalit2017TARNET}
Shalit, U., et~al.: Estimating individual treatment effect: generalization bounds and algorithms. PMLR  (2017)

\bibitem{SerstinskyRNN}
Sherstinsky, A.: Fundamentals of recurrent neural network {(RNN)} and long short-term memory {(LSTM)} network. CoRR  (2018)

\bibitem{Tang2022}
Tang, W., et~al.: Soden: A scalable continuous-time survival model through ordinary differential equation networks. The Journal of Machine Learning Research  (2022)

\bibitem{tousignant2019}
Tousignant, A., et~al.: Prediction of disease progression in multiple sclerosis patients using deep learning analysis of mri data. In: Proceedings of The 2nd International Conference on Medical Imaging with Deep Learning. PMLR (2019)

\bibitem{Verhelst2023}
Verhelst, T., et~al.: Uplift vs. predictive modeling: a theoretical analysis. arXiv  (2023)

\bibitem{BRAVO}
Vollmer, T.L., et~al.: {{A} randomized placebo-controlled phase {I}{I}{I} trial of oral laquinimod for multiple sclerosis}. J Neurology  (2014)

\bibitem{ma2023treatmentHemmorage}
Wenao, M., et~al.: Treatment outcome prediction for intracerebral hemorrhage via generative prognostic model with imaging and tabular data (2023)

\bibitem{Xuechen2020}
Xuechen, L., et~al.: Scalable gradients for stochastic differential equations. CoRR  (2020)

\end{thebibliography}
\end{document}